# Adaptive Boosting with Fairness-aware Reweighting Technique for Fair Classification

Xiaobin Song[1*], Zeyuan Liu[1*], Benben Jiang[1,2]†

1 Department of Automation, Tsinghua University, Beijing 100084, China

2 Beijing National Research Center for Information Science and Technology,

Tsinghua University, Beijing 100084, China

## Abstract

Machine learning methods based on AdaBoost have been widely applied to various classification problems across many mission-critical applications including healthcare, law and finance. However, there is a growing concern about the unfairness and discrimination of data-driven classification models, which is inevitable for classical algorithms including AdaBoost. In order to achieve fair classification, a novel fair AdaBoost (FAB) approach is proposed that is an interpretable fairness-improving variant of AdaBoost. We mainly investigate binary classification problems and focus on the fairness of three different indicators (i.e., accuracy, false positive rate and false negative rate). By utilizing a fairness-aware reweighting technique for base classifiers, the proposed FAB approach can achieve fair classification while maintaining the advantage of AdaBoost with negligible sacrifice of predictive performance. In addition, a hyperparameter is introduced in FAB to show preferences for the fairness-accuracy trade-off. An upper bound for the target loss function that quantifies error rate and unfairness is theoretically derived for FAB, which provides a strict theoretical support for the fairness-improving methods designed for AdaBoost. The effectiveness of the proposed method is demonstrated on three real-world datasets (i.e., Adult, COMPAS and HSLS) with respect to the three fairness indicators. The results are accordant with theoretic analyses, and show that (i) FAB significantly improves classification fairness at a small cost of accuracy compared with AdaBoost; and (ii) FAB outperforms state-of-the-art fair classification methods including equalized odds method, exponentiated gradient method, and disparate mistreatment method in terms of the fairness-accuracy trade-off.

---

*These authors contributed equally to this work
†Corresponding author: bbjiang@tsinghua.edu.cn

## 1. Introduction

Machine learning has been widely applied to a variety of practical decision-making problems and achieved great success (Anahideh *et al.*, 2022; Bertolini *et al.*, 2021; Jiang, 2021). However, in many real-world cases such as loan approvals (Mukerjee *et al.*, 2002) and recidivism prediction (Zeng *et al.*, 2017), people not only focus on the prediction accuracy, but also consider decision fairness. The underlying reason for fairness concern is that general classification models may unfairly treat people with different sensitive attributes (e.g., gender and race), which may cause great social problems and should be avoided in practical decision-making processes. For example, a common crime assessment tool, COMPAS, has been proved to be racially biased (Dressel & Farid, 2018). It underestimates the recidivism rate of white defendants and overestimates that of black defendants, which cannot be drawn from slight differences in the prediction accuracy between two groups of defendants. Another well-known example is an unfair resume screening algorithm about gender (Zehlike *et al*., 2017). In an online job platform named XING, the proportion of two genders among top candidates is largely unbalanced, which is unfavorable for capable women applicants (Lahoti *et al*., 2019). From the aforementioned cases, the main challenges of machine learning based decision-making models can be summarized as follows: (1) The machine learning based models may render unfair decisions among vulnerable groups because of biased datasets over sensitive attributes. (2) Even though some datasets do not contain sensitive attributes, other features in the datasets may have hidden correlations with sensitive attributes, leading to bias in decision making. Hence, it's important to research for new methods to improve the fairness of classification methods.

Fairness-improving methods vary in the notions of targeted fairness and the mechanisms used. Based on the stage when fairness-improving mechanisms employed, methods that enhance the algorithmic fairness can be categorized into three groups: pre-processing, in-processing, and post-processing (Islam *et al*., 2022; Mehrabi *et al*., 2021). Pre-processing methods chiefly aim to mitigate the unfairness by processing the input data before training data-driven models, motivated by the fact that data-driven algorithms reflect biases and trends in training data (Islam *et al*., 2022). This process is typically realized by methods such as generating new representations of features without correlations with protected attributes, or modifying distributions of labels according to relevant fairness notions (Fish *et al*., 2016; Kamiran & Calders, 2012). For example, Kamiran and Calders (2012) proposed a fairness-improving method by resampling the input data with a weighted sampling method to guarantee the statistical independence between ground-truth labels and sensitive attributes. Another method enforcing independence between labels and sensitive attributes was proposed in Feldman *et al*. (2015), which alters the values of each attribute until the marginal distributions are similar between the favored and unfavored groups to achieve demographic parity. Zhang *et al*. (2017) developed a causal model-based method to discover corresponding causal association between the sensitive attributes and labels, and alter the labels to remove such causal influences, which can achieve path-specific fairness. Salimi *et al*. (2019) put forward a pre-processing method to enforce justifiable fairness, which prohibited causal association between the labels and sensitive attributes. As a model-agnostic method, pre-processing enjoys the main advantage of flexibility in choosing the algorithms according to different application scenarios. Moreover, as pre-processing handles the data before training to remove biases, this method can subsequently guarantee that the target fairness notion is satisfied by trained models. However, because pre-processing takes effect before training, these approaches can only support a limited range of fairness notions and may not provably ensure the algorithmic fairness.

In-processing methods aim to improve the fairness of learning process itself, which are most favored by machine learning researchers (Kearns *et al*., 2018; Zhang *et al*., 2018). In-processing methods work during the training process and fairness is typically achieved by approaches such as adopting fairness constraints (Zafar *et al*., 2017; Agarwal *et al*., 2018), using a unfairness-penalizing loss function (Iosifidis & Ntoutsi, 2019), or training with an adversary that improves the model performance in the protected-group (Grari *et al*., 2019). For example, Zafar *et al*. (2017) developed a method to enforce equalized odds by translating the fairness notion to a convex function of classifier parameters, and solving the optimization problem that maximize prediction accuracy under fairness constraints. To enforce equalized odds, Zhang *et al*. (2018) proposed an adversarial learning-based approach to obtain optimal classifiers such that the predicted labels has no information about sensitive attributes. Lee *et al.* (2021) developed a support vector method-based method for fair clustering by achieving fair correspondence distribution for each cluster. Celis *et al.* (2019) proposed a method to accommodate multiple fairness notions in a general framework by solving a linear constrained optimization problem to improve accuracy under fairness constraints. Thomas *et al.* (2019) developed a framework to unify a variety of notions, which ensured that the trained model can satisfy the fairness notions with high probability by computing optimal classifier parameters for which the maximum possible fairness violation is within certain threshold. Because in-processing methods work within the model-training stage, they enjoy the advantage of achieving fairness requirements directly through designing classification objectives. Therefore, they have the potential to provide guarantees for fairness.

Post-processing methods aim to mitigate unfairness by manipulating the outputs of models to satisfy statistical measures corresponding to different fairness notions. Post-processing methods take effect during output stages and fairness is typically improved through techniques including modifying the decision boundary of particular sub-groups (Hardt *et al*., 2016), or randomly classifying a portion of disadvantaged groups (Pleiss *et al*., 2017). For example, Kamiran *et al.* (2012) enforced demographic parity by deriving a critical region around decision boundaries and randomly modifying outputs in that region until the results have similar probability across sensitive groups. Another method enforcing equalized odds was proposed in Hardt *et al.* (2016) by learning the parameters of a mapping from original output to modified predictions. Pleiss *et al.* (2017) enforced predictive equality or equal opportunity by modifying outputs for a random subset of tuples until their chosen fairness metric is equalized. Nguyen *et al.* (2021) developed a Gaussian process-based method for fair classification, which formulated relabeling process as an optimization problem to maximize the fairness and minimize the difference between the pre-trained and relabeled classifiers. Karmiran *et al.* (2018) developed a social discrimination control method for fair classification, by exploiting the decision theoretical notion of reject option to handle the instances with uncertain labels. Because post-processing takes effect within the output stage, they enjoy the benefits as model-agnostic methods being compatible with score-based classifiers. However, as post-processing is adopted in the last stage during the learning process, it shows less flexibility on fair classification performance than pre-processing and in-processing methods.

Among the aforementioned three main directions for fair classification, in-processing method has better performance and robustness since it directly mitigates unfairness in model training process (Kearns *et al*., 2018; Zhang *et al*., 2018). It is therefore more favored by machine learning researchers, and a variety of fair-improvement algorithms for existing classification methods (e.g., convex-margin-based classifiers) have been put forward. However, in-processing approaches are model-specific, and most of the methods are not suitable for boosting algorithms (Huang *et al.*, 2022).

As a widely used boosting classifier, AdaBoost shows good performance on interpretability and accuracy (Freund & Schapire, 1996). Accuracy-improving techniques about AdaBoost has been developed over years (Hastie *et al.*, 2009; Li *et al.*, 2008). For example, a support vector machine (SVM) based AdaBoost was proposed in Li *et al.* (2008) to balance the classification accuracy and diversity. Wang and Sun (2021) developed an improved version of AdaBoost to enhance the classification performance on imbalanced datasets. A variant of AdaBoost was developed in Hastie *et al.* (2009) that extends the original AdaBoost algorithm to a multi-class version, which showed competitive performance in multi-classification tasks. Along with classification accuracy, fairness is another important concern among the machine learning community, and some fairness-improving methods tailored for AdaBoost have been proposed. For instance, Huang *et al.* (2022) put forward an AdaBoost based algorithm to achieve fair binary classification, considering fairness notions of average odds difference and equal opportunity difference. However, most of previous works on improving the fairness of AdaBoost only validated their effectiveness through empirical performance. Few of them demonstrated their classification ability, scalability and potential of improvement from a theoretical perspective. In this regard, we develop a novel fair adaptive boosting (FAB) approach, an interpretable fairness-improving variant of AdaBoost to address this unsolved issue.

AdaBoost is a practical boosting classifier that iteratively calls weak learners for training according to their accuracy-based weights. In this work, we mainly concentrate on updating the weights in a fairness-aware manner to achieve better performance in fair classification. In terms of fairness, we consider three different indicators (i.e., accuracy, false positive rate (FPR) and false negative rate (FNR)) between the favored and unfavored groups. To improve fairness while maintaining accuracy, we add fairness loss to the target loss function of FAB. A hyperparameter $\lambda$ is introduced into the target loss function to represent the preference for fairness over accuracy, and a new solution is derived for the proposed FAB method. More importantly, we theoretically provide an upper bound for the target loss function of fairness-accuracy to enhance the interpretability of the proposed method, which can guarantee the fairness of the FAB approach. The optimization objective is then altered from the target loss function to its upper bound, and the optimal update algorithm for sample weights is then derived using greedy algorithm.

The main contributions of our work are summarized as:

1. A fair AdaBoost approach is developed that accounts for both accuracy and fairness. Three widely used indicators (accuracy, FPR and FNR) is used to quantify fairness, and a hyperparameter is set to show preferences for the trade-off between classification accuracy and fairness.

2. An upper bound is theoretically derived for the target loss function of FAB that represents the trade-off between accuracy and the fairness of an indicator. An efficient computational algorithm based on greedy technique is then designed to optimize this upper bound instead of the target function. To the best of the authors' knowledge, it is the first time to provide strict theoretical support for the fairness-improving methods designed for AdaBoost.

3. We evaluate the proposed approach on three real-world datasets, covering all three fairness indicators. The results show that (i) our FAB approach can largely enhance fairness at the slight expense of accuracy in most cases, and (ii) FAB possesses better performance than state-of-the-art baselines including equalized odds method, exponentiated gradient method, and disparate mistreatment method.

The remainder of the article is organized as follows. Section 2 describes the overall framework and implementation details of the proposed FAB approach. The theoretical analysis of the proposed approach is provided in Section 3, in which an upper bound on the target loss function is obtained. In Section 4, the effectiveness of the proposed method is demonstrated on three real-world datasets, followed by the conclusions in Section 5.

## 2. Methodology

### 2.1. Problem definition

In this article we mainly discuss binary classification problems and focus on the fairness of three different indicators: accuracy, false positive rate (FPR) and false negative rate (FNR). Suppose $f(x)$ is a classifier (e.g., decision tree) trained on a dataset $\mathcal{D} = \{x_i, y_i\}_{i=1}^{N}$, where $x_i$ represents a set of features and $y_i \in \{-1, +1\}$ represents a binary true label. In $x_i$, there is a sensitive feature $S_i \in \{0, 1\}$. Let $S_i = 1$ represent the favored group, which typically has higher accuracy, lower FPR and FNR, compared with the unfavored group represented by $S_i = 0$. $f(x)$ is expected to predict a label $\hat{y}_i \in \{-1, +1\}$ according to $x_i$. Referring to the concept of disparate mistreatment proposed by Zafar (2019), our definitions of fairness loss can be demonstrated as follows:

**Definition 1.** (Fairness loss of accuracy). We define fairness loss of accuracy as $|P(f(x) = y|S = 1) - P(f(x) = y|S = 0)|$, which represents the difference between the accuracy of the favored and unfavored groups.

**Definition 2.** (Fairness loss of FPR). We define fairness loss of FPR as $|P(f(x) \neq y|S = 1, y = -1) - P(f(x) \neq y|S = 0, y = -1)|$, which represents the difference between the false positive rate of the favored and unfavored groups.

**Definition 3.** (Fairness loss of FNR). We define fairness loss of FNR as $|P(f(x) \neq y|S = 1, y = +1) - P(f(x) \neq y|S = 0, y = +1)|$, which represents the difference between the false negative rate of the favored and unfavored groups.

**Problem statement.** When considering the fairness of accuracy, we assume that for general classifiers trained on a training set $\mathcal{D} = \{x_i, y_i\}_{i=1}^N$, the classification accuracy of the favored group is higher than that of the unfavored group: $P(f(x) = y|S = 1) > P(f(x) = y|S = 0)$ (i.e. $P(f(x) \neq y|S = 1) < P(f(x) \neq y|S = 0)$). Our assumption is reasonable since traditional classification methods only focus on accuracy and rarely consider fairness. To achieve higher accuracy, general classifiers may form preference for the accuracy of favored groups through the training process unintentionally. Similarly, we assume that $P(f(x) \neq y|S = 1, y = -1) < P(f(x) \neq y|S = 0, y = -1)$ when considering the fairness of FPR, and $P(f(x) \neq y | S = 1, y = +1) < P(f(x) \neq y | S = 0, y = +1)$ when considering the fairness of FNR. Based on the aforementioned fairness notions, our objective is to construct a classifier $f(x)$ that achieves both high accuracy and low fairness loss.

*2.2. Fair adaptive boosting (FAB)*

Let $\{h_t(x), \alpha_t\}_{t=1}^{T}$ be a set of base classifiers and their weights, $f(x)$ be the ensemble classifier. That is:

$$f(x) = \text{sign}(\sum_{t=1}^{T} \alpha_t h_t(x)) \qquad (1)$$

Take the fairness of accuracy as example. Let $D_t$ be the sample weight in the $t$th iteration. Since basic AdaBoost is designed to perform well in accuracy only, to improve the fairness score of $f(x)$, we introduce fairness loss to the loss function. Hence, our method can be formalized as an optimization problem corresponding to the following objective function (Jeong *et al.*, 2022):

$$\underset{f(x)}{\text{argmin}} \frac{1}{N}\sum_{i=1}^{N} \mathbb{I}(f(x_i) \neq y_i) + \lambda \left| \frac{\sum_{i=1}^{N} \mathbb{I}(S_i=0)\mathbb{I}(f(x_i) \neq y_i)}{\sum_{i=1}^{N} \mathbb{I}(S_i=0)} - \frac{\sum_{i=1}^{N} \mathbb{I}(S_i=1)\mathbb{I}(f(x_i) \neq y_i)}{\sum_{i=1}^{N} \mathbb{I}(S_i=1)} \right| \qquad (2)$$

where the notation $\mathbb{I}(\partial)$ is defined as:

$$\mathbb{I}(\partial) = \begin{cases} 1, & \text{if and only if } \partial \text{ is true} \\ 0, & \text{otherwise} \end{cases} \qquad (3)$$

It is worth mentioning that the proposed loss function comprises two interpretable terms. The first term means the rate of misclassified samples (error rate) and the second term means the difference between the error rates of favored and unfavored groups. We introduce the hyperparameter $\lambda \in \left[0, \frac{\sum_{i=1}^{N} \mathbb{I}(S_i=1)}{N}\right]$ to represent the trade-off between accuracy and the fairness of accuracy. For a larger value of $\lambda$, the classification model would emphasize more on fairness performance but less on accuracy, and vice versa.

To simplify the representation of Eq. (2), we define $D_1$ as:

$$D_1^{acc}(x_i) = \begin{cases} \frac{1}{N} + \frac{\lambda}{\sum_{j=1}^N \mathbb{I}(S_j=0)}, & if\ S_i = 0 \\ \frac{1}{N} - \frac{\lambda}{\sum_{j=1}^N \mathbb{I}(S_j=1)}, & if\ S_i = 1 \end{cases} \quad (4)$$

According to the AdaBoost algorithm, the update method of base classifier weight and sample weight can be obtained as:

$$e_t = \sum_{j=1}^N D_t(x_j)\mathbb{I}(h_t(x_j) \neq y_j) \quad (5)$$

$$\alpha_t = \frac{1}{2}\ln\left(\frac{1-e_t}{e_t}\right) \quad (6)$$

$$D_{t+1}(x_i) = \frac{D_t(x_i)e^{-y_i\alpha_t h_t(x_i)}}{\sum_{j=1}^N D_t(x_j)e^{-y_j\alpha_t h_t(x_j)}} \triangleq D_t(x_i)\frac{e^{-y_i\alpha_t h_t(x_i)}}{Z_t(h_t,\alpha_t)} \quad (7)$$

where $Z_t(h_t,\alpha_t) = \sum_{j=1}^N D_t(x_j)e^{-y_j\alpha_t h_t(x_j)}$ is a normalization factor. Note that $(D_t, Z_t)$ can refer to $(D_t^{acc}, Z_t^{acc})$, $(D_t^{FPR}, Z_t^{FPR})$, and $(D_t^{FNR}, Z_t^{FNR})$ in Eq. (5) and (7). This representation form is also used in Section 4 for brevity. Noticeably, when training the base classifiers on the training set with $D_t$, sensitive features $S_i$ are not used for fairness concerns.

We refer to the aforementioned process as fair adaptive boosting (FAB), which is summarized in Algorithms 1 and 2. It is worth mentioning that FAB utilizes the framework of AdaBoost and update the weight $D_1$ to enhance the fairness. The ensemble classifier $f(x)$ after training is therefore referred to as a fair classifier.

Similarly, in terms of the fairness of FPR, our proposed objective function can be formulated as:

$$\underset{f(x)}{\operatorname{argmin}} \frac{1}{N}\sum_{i=1}^N \mathbb{I}(f(x_i) \neq y_i) + \lambda \left| \frac{\sum_{i=1}^N \mathbb{I}(S_i=0)\mathbb{I}(y_i=-1)\mathbb{I}(f(x_i)\neq y_i)}{\sum_{i=1}^N \mathbb{I}(S_i=0)\mathbb{I}(y_i=-1)} - \frac{\sum_{i=1}^N \mathbb{I}(S_i=1)\mathbb{I}(y_i=-1)\mathbb{I}(f(x_i)\neq y_i)}{\sum_{i=1}^N \mathbb{I}(S_i=1)\mathbb{I}(y_i=-1)} \right| \quad (8)$$

where $\lambda \in \left[0, \frac{\sum_{i=1}^{N} \mathbb{I}(S_i=1)\mathbb{I}(y_i=-1)}{N}\right]$ is the hyperparameter for the trade-off between accuracy and the fairness of FPR. The initial sample weight $D_1$ can be modified into the following form with respect to FPR:

$$D_1^{FPR}(x_i) = \begin{cases} \frac{1}{N} + \frac{\lambda}{\sum_{j=1}^{N} \mathbb{I}(S_j=0)\mathbb{I}(y_j=-1)}, & \text{if } S_i = 0 \text{ and } y = -1 \\ \frac{1}{N} - \frac{\lambda}{\sum_{j=1}^{N} \mathbb{I}(S_j=1)\mathbb{I}(y_j=-1)}, & \text{if } S_i = 1 \text{ and } y = -1 \\ \frac{1}{N} & otherwise \end{cases} \quad (9)$$

When considering FNR, the objective function of our optimization problem can be described as:

$$\operatorname*{argmin}_{f(x)} \frac{1}{N}\sum_{i=1}^{N} \mathbb{I}(f(x_i) \neq y_i) + \lambda \left| \frac{\sum_{i=1}^{N} \mathbb{I}(S_i=0)\mathbb{I}(y_i=+1)\mathbb{I}(f(x_i)\neq y_i)}{\sum_{i=1}^{N} \mathbb{I}(S_i=0)\mathbb{I}(y_i=+1)} - \frac{\sum_{i=1}^{N} \mathbb{I}(S_i=1)\mathbb{I}(y_i=+1)\mathbb{I}(f(x_i)\neq y_i)}{\sum_{i=1}^{N} \mathbb{I}(S_i=1)\mathbb{I}(y_i=+1)} \right| \quad (10)$$

where $\lambda \in \left[0, \frac{\sum_{i=1}^{N} \mathbb{I}(S_i=1)\mathbb{I}(y_i=+1)}{N}\right]$ is the hyperparameter for the trade-off between accuracy and the fairness of FNR. The initial sample weight $D_1$ should be modified into the following form accordingly:

$$D_1^{FNR}(x_i) = \begin{cases} \frac{1}{N} + \frac{\lambda}{\sum_{j=1}^{N} \mathbb{I}(S_j=0)\mathbb{I}(y_j=+1)}, & \text{if } S_i = 0 \text{ and } y = +1 \\ \frac{1}{N} - \frac{\lambda}{\sum_{j=1}^{N} \mathbb{I}(S_j=1)\mathbb{I}(y_j=+1)}, & \text{if } S_i = 1 \text{ and } y = +1 \\ \frac{1}{N} & otherwise \end{cases} \quad (11)$$

The FAB algorithm for the fairness indicators of accuracy, FPR and FNR can be summarized as Algorithms 1 and 2, respectively.

**Algorithm 1:** The proposed FAB algorithm for the fairness indicator of accuracy

**Input:** $\mathcal{D} = \{x_i, y_i\}_{i=1}^{N}$: training set with sensitive features

$\lambda \in \left[0, \frac{\sum_{i=1}^{N} \mathbb{I}(S_i=1)}{N}\right]$: hyper-parameter for a trade-off between classification error and the fairness of accuracy

$T$: # of iterations

**begin**

$$D_1^{acc}(x_i) = \begin{cases} \frac{1}{N} + \frac{\lambda}{\sum_{j=1}^{N} \mathbb{I}(S_j=0)}, & \text{if } S_i = 0 \\ \frac{1}{N} - \frac{\lambda}{\sum_{j=1}^{N} \mathbb{I}(S_j=1)}, & \text{if } S_i = 1 \end{cases} \quad i = 1, 2, \cdots, N$$

for $t = 1, 2, \cdots, T$

train $h_t$ on $\mathcal{D}$ with the sample weight $D_t^{acc}$ (do not use sensitive features $S_i$)

$e_t = \sum_{j=1}^{N} D_t^{acc}(x_j) \mathbb{I}(h_t(x_j) \neq y_j)$

$\alpha_t = \frac{1}{2} \ln\left(\frac{1-e_t}{e_t}\right)$

$D_{t+1}^{acc}(x_i) = \frac{D_t^{acc}(x_i) e^{-y_i \alpha_t h_t(x_i)}}{\sum_{j=1}^{N} D_t^{acc}(x_j) e^{-y_j \alpha_t h_t(x_j)}} \quad i = 1, 2, \cdots, N$

**end for**

**Output:** $f(x) = \text{sign}\left(\sum_{t=1}^{T} \alpha_t h_t(x)\right)$

**end**

---

**Algorithm 2:** The proposed FAB algorithm for the fairness indicator of FPR

**Input:** $\mathcal{D} = \{x_i, y_i\}_{i=1}^{N}$: training set with sensitive features

$\lambda \in \left[0, \frac{\sum_{i=1}^{N} \mathbb{I}(S_i=1)\mathbb{I}(y_i=-1)}{N}\right]$: hyper-parameter for a trade-off between classification error and the fairness of FPR

$T$: # of iterations

**begin**

$$D_1^{FPR}(x_i) = \begin{cases} \frac{1}{N} + \frac{\lambda}{\sum_{j=1}^{N} \mathbb{I}(S_j=0)\mathbb{I}(y_j=-1)}, & \text{if } S_i = 0 \text{ and } y = -1 \\ \frac{1}{N} - \frac{\lambda}{\sum_{j=1}^{N} \mathbb{I}(S_j=1)\mathbb{I}(y_j=-1)}, & \text{if } S_i = 1 \text{ and } y = -1 \quad i = 1, 2, \cdots, N \\ \frac{1}{N} & \text{otherwise} \end{cases}$$

for $t = 1, 2, \cdots, T$

train $h_t$ on $\mathcal{D}$ with the sample weight $D_t^{FPR}$ (do not use sensitive features $S_i$)

$e_t = \sum_{j=1}^{N} D_t^{FPR}(x_j) \mathbb{I}(h_t(x_j) \neq y_j)$

$\alpha_t = \frac{1}{2} \ln\left(\frac{1-e_t}{e_t}\right)$

$D_{t+1}^{FPR}(x_i) = \frac{D_t^{FPR}(x_i) e^{-y_i \alpha_t h_t(x_i)}}{\sum_{j=1}^{N} D_t^{FPR}(x_j) e^{-y_j \alpha_t h_t(x_j)}} \quad i = 1, 2, \cdots, N$

**end for**

**Output:** $f(x) = \text{sign}\left(\sum_{t=1}^{T} \alpha_t h_t(x)\right)$

**end**

The proposed FAB algorithm for the fairness indicator of FNR can be obtained by replacing the terms of $\mathbb{I}(y_i = -1)$ and $y = -1$ in Algorithm 2 as $\mathbb{I}(y_i = +1)$ and $y = +1$, respectively. Then the sample weight $D_t^{FNR}$ can be calculated following the same way for $D_t^{FPR}$ in Algorithm 2. The detailed implementation for the FNR fairness is omitted for brevity.

**3. Theoretical analysis**

In this section, the theoretical analysis of FAB will be conducted. We select exponential loss function and provide upper bounds on the Eq. (2) (accuracy), Eq. (8) (FPR) and Eq. (10) (FNR), respectively. The detailed proof is provided below.

*3.1. Upper bound on the objective loss function*

Before formally deriving upper bound, we focus on the update technique of sample weights first. Eq. (7) can be rewritten as:

$$D_{t+1}(x_i) = D_t(x_i) \frac{e^{-y_i \alpha_t h_t(x_i)}}{Z_t} = D_{t-1}(x_i) \frac{e^{-y_i(\alpha_{t-1} h_{t-1}(x_i) + \alpha_t h_t(x_i))}}{Z_{t-1} Z_t} = \cdots = D_1(x_i) \frac{e^{-y_i \sum_{j=1}^t \alpha_j h_j(x_i)}}{\prod_{j=1}^t Z_j} \quad (12)$$

Considering the normalization factor $Z_t$ in Eq. (7), it can be obtained as:

$$\sum_{i=1}^N D_t(x_i) = 1 \quad (13)$$

Combining this with Eq. (12), we obtain:

$$\sum_{i=1}^N D_1(x_i) e^{-y_i \sum_{j=1}^t \alpha_j h_j(x_i)} = \prod_{j=1}^t Z_j(h_j, \alpha_j) \quad (14)$$

*3.1.1. Upper bound for the fairness of accuracy*

In this section, we will derive the upper bound for our objective loss function for the fairness of accuracy (i.e., Eq. (2)). Given the assumption mentioned in problem statement (i.e., for general classifiers, there is $P(f(x) \neq y | S = 1) <$

$P(f(x) \neq y | S = 0))$, the objective loss function can be rewritten as:

$$L_{acc}(f(x)) = \frac{1}{N}\sum_{i=1}^{N} \mathbb{I}(f(x_i) \neq y_i) + \lambda \left( \frac{\sum_{i=1}^{N} \mathbb{I}(S_i=0)\mathbb{I}(f(x_i) \neq y_i)}{\sum_{i=1}^{N} \mathbb{I}(S_i=0)} - \frac{\sum_{i=1}^{N} \mathbb{I}(S_i=1)\mathbb{I}(f(x_i) \neq y_i)}{\sum_{i=1}^{N} \mathbb{I}(S_i=1)} \right) = \sum_{i=1}^{N} \mathbb{I}(S_i = 0) \left( \frac{1}{N} + \frac{\lambda}{\sum_{i=1}^{N} \mathbb{I}(S_i=0)} \right) \mathbb{I}(f(x_i) \neq y_i) + \sum_{i=1}^{N} \mathbb{I}(S_i = 1) \left( \frac{1}{N} - \frac{\lambda}{\sum_{i=1}^{N} \mathbb{I}(S_i=1)} \right) \mathbb{I}(f(x_i) \neq y_i) \quad (15)$$

We can obtain the following relationship between 0/1 loss and exponential loss:

$$\mathbb{I}(f(x_i) \neq y_i) \leq e^{-\alpha y_i f(x_i)} \quad \forall 1 \leq i \leq n, \alpha > 0 \quad (16)$$

Combining this with Eq. (1), Eq. (4) and Eq. (14), when $\lambda \in \left[0, \frac{\sum_{i=1}^{N} \mathbb{I}(S_i=1)}{N}\right]$, an upper bound for the objective loss function can be derived:

$$L_{acc}(f(x)) \leq \sum_{i=1}^{N} \mathbb{I}(S_i = 0) \left( \frac{1}{N} + \frac{\lambda}{\sum_{i=1}^{N} \mathbb{I}(S_i=0)} \right) e^{-y_i \sum_{t=1}^{T} \alpha_t h_t(x_i)} + \sum_{i=1}^{N} \mathbb{I}(S_i = 1) \left( \frac{1}{N} - \frac{\lambda}{\sum_{i=1}^{N} \mathbb{I}(S_i=1)} \right) e^{-y_i \sum_{t=1}^{T} \alpha_t h_t(x_i)} = \sum_{i=1}^{N} D_1^{acc}(x_i) e^{-y_i \sum_{t=1}^{T} \alpha_t h_t(x_i)} = \prod_{t=1}^{T} Z_t^{acc}(h_t, \alpha_t) \quad (17)$$

*3.1.2. Upper bound for the fairness of FPR and FNR*

Similar to the derivation presented above, for FPR, given the assumption mentioned in problem statement (i.e., for general classifiers, there is $P(f(x) \neq y | S = 1, y = -1) < P(f(x) \neq y | S = 0, y = -1)$), the following upper bound for Eq. (8) when $\lambda \in \left[0, \frac{\sum_{i=1}^{N} \mathbb{I}(S_i=1)\mathbb{I}(y_i=-1)}{N}\right]$ can be derived:

$$L_{FPR}(f(x)) \leq \prod_{t=1}^{T} Z_t^{FPR}(h_t, \alpha_t) \quad (18)$$

For FNR, given the assumption mentioned in problem statement (i.e., for general classifiers, there is $P(f(x) \neq y | S = 1, y = +1) < P(f(x) \neq y | S = 0, y = +1)$), when $\lambda \in \left[0, \frac{\sum_{i=1}^{N} \mathbb{I}(S_i=1)\mathbb{I}(y_i=+1)}{N}\right]$, the following upper bound for Eq. (10) can be obtained:

$$L_{FNR}(f(x)) \leq \prod_{t=1}^{T} Z_t^{FNR}(h_t, \alpha_t) \quad (19)$$

The detailed proofs on upper bound for the fairness of FPR and FNR are provided in the section of Appendix.

The aforementioned results show that our objective loss functions have common upper bounds, namely, the product of normalize factors $\{Z_t\}_{t=1}^T$. Our proofs also interpret why we modify $D_1$ to be Eq. (4), Eq. (9) and Eq. (11) respectively.

*3.2. Calculation of $\alpha_t$*

Inspired by AdaBoost, we focus on the upper bound in Eq. (17), Eq. (18) and Eq. (19). Instead of directly optimizing Eq. (2), Eq. (8) and Eq. (10), we optimize their upper bounds. The objective function can be stated as:

$$\underset{\{h_t,\alpha_t\}_{t=1}^T}{\mathrm{argmin}} \prod_{t=1}^T Z_t(h_t, \alpha_t) \qquad (20)$$

By adopting the greedy strategy, the optimization problem can be rewritten as:

$$(h_t, \alpha_t)^* = \underset{(h_t,\alpha_t)}{\mathrm{argmin}} Z_t(h_t, \alpha_t) = \underset{(h_t,\alpha_t)}{\mathrm{argmin}} \sum_{j=1}^N D_t(x_j) e^{-y_j \alpha_t h_t(x_j)} \qquad (21)$$

By computing the partial derivative of $Z_t(h_t, \alpha_t)$ with respect to $\alpha_t$, and letting it equal to zero, the value of $\alpha_t$ where Eq. (21) reaches its minimum can be derived as:

$$\alpha_t = \frac{1}{2} \ln\left(\frac{1-e_t}{e_t}\right) \qquad (22)$$

**4. Results**

In this section, we evaluate the performance of the proposed FAB approach on three real-world datasets (Adult, Compas and HSLS). AdaBoost, equalized odds method (Hardt *et al*., 2016), exponentiated gradient method (Agarwal *et al*., 2018), and disparate mistreatment method (Zafar *et al*., 2019) are chosen as baselines to see how the FAB method can improve fairness score while maintaining accuracy.

*4.1. Datasets*

We use three datasets, Adult, Compas and HSLS, in our experiment, which are widely used to test fair classification methods.

1. Adult (Nguyen *et al*., 2021): Adult is a dataset about income, from which we use 14 features in our experiment. Among them, we choose *gender* as the sensitive feature and *income* as the label (Jeong *et al*., 2022). For Adult, we choose two different fairness indicators: accuracy and FPR.

2. Compas (Nguyen *et al*., 2021): Compas is a credit scoring dataset. We use eight features from Compas, where *race* is chosen as the sensitive feature and *two_year_recid* is chosen as the binary label in our experiment (Jeong *et al*., 2022). We choose FNR as our fairness indicator.

3. HSLS (Jeong *et al*., 2022): HSLS is a dataset about the math test performance of high school students. We select 11 features from thousands of features in the original dataset and choose *race* as a sensitive feature (Jeong *et al*., 2022). The binary label is created from *X1TXMSCR* (continuous test score). Since our method does not consider the situation of missing data, we delete all samples with missing values. For HSLS, we choose FNR as the fairness indicator.

In Section 3, a value range for $\lambda$ is used (i.e., $\lambda \in \left[0, \frac{\sum_{i=1}^{N} \mathbb{I}(S_i=1)}{N}\right]$ for the fairness of accuracy, $\lambda \in \left[0, \frac{\sum_{i=1}^{N} \mathbb{I}(S_i=1)\mathbb{I}(y_i=-1)}{N}\right]$ for FPR and $\lambda \in \left[0, \frac{\sum_{i=1}^{N} \mathbb{I}(S_i=1)\mathbb{I}(y_i=+1)}{N}\right]$ for FNR). The value range of $\lambda$ is related to the proportion of the favored group. However, two groups in the original dataset are often imbalanced, and low proportion of favored group may narrow the value range of $\lambda$. Hence, we balance all three datasets in terms of the sensitive feature. Moreover, we also balance the Adult dataset with labels of 0 and 1 due to the high imbalance of the original data (Jeong *et al*., 2022). After balancing, we have 7834 data points for Adult, 4206 data points for Compas and 8414 data points for HSLS.

*4.2. Baselines*

The proposed FAB fair classification approach is evaluated in comparison with four baselines, namely, the classical AdaBoost (Freund & Schapire, 1996) and the state-of-the-art fair classification methods of equalized odds method (Hardt *et al*., 2016), exponentiated gradient method (Agarwal *et al*., 2018), and disparate mistreatment method (Zafar *et al*., 2019) (we use the name of their first authors, Hardt, Agarwal, and Zafar, to refer to them respectively). We select the classical AdaBoost to see how much we can improve fairness while maintaining accuracy, and choose Hardt, Agarwal, and Zafar methods to see how much our approach can improve the performance of balancing accuracy and fairness.

*4.3. Implementation details*

In this section, the details of experimental implementation is provided. For FAB, AdaBoost, Hardt and Agarwal methods, CART Decision Tree is selected as base classifiers using the Python Scikit-learn module. The maximum depth of decision trees is set as 3. From Agarwal *et al*. (2018), the hyperparameter $\varepsilon$ is set as {0.001, 0.005, 0.01, 0.05, 0.1}. Other hyperparameters are summarized in Table 1.

Table 1 Hyperparameters for the proposed FAB method and baselines.

| Dataset | Fairness indicator | AdaBoost $T$ | FAB $T$ | FAB $\lambda$ | Zafar $\tau$ |
|---|---|---|---|---|---|
| Adult | accuracy | 30 | 30 | 0.1, 0.2, 0.3, 0.4, 0.45, 0.5 | 0.5, 1, 5, 10, 100 |
| Adult | FPR | 30 | 30 | 0.1, 0.15, 0.2, 0.25, 0.3 | 0.01, 0.1, 1, 10, 100 |
| Compas | FNR | 30 | 30 | 0.1, 0.2, 0.3, 0.35, 0.4, 0.45 | 0.01, 0.1, 1, 10, 100 |
| HSLS | FNR | 20 | 20 | 0.05, 0.1, 0.15, 0.2, 0.25 | 0.1, 1, 10, 100 |

For the model training step, 70% of the dataset is used, and the remaining 30% is used for model testing. Each set of hyperparameters use 20 random seeds to improve the confidence of experiment results.

*4.4. Performance comparison*

Our experiment illustrates the performance of the proposed FAB approach and baselines on both training set and testing set. The results are depicted in Figs. 1, 2, 3, 4 and some representative results for the testing set are displayed in Table 2. Compared with classical AdaBoost, when considering the fairness of accuracy and FPR for Adult, our FAB method can reduce the fairness loss by 83.5% (from 0.08 to 0.01) and 88.4% (from 0.21 to 0.02) at the cost of sacrificing the accuracy by 4.4% (from 0.83 to 0.79) and 2.4% (from 0.83 to 0.81) respectively for the testing set. When it comes to FNR for Compas and HSLS, the proposed FAB method can achieve the effect of reducing the fairness loss by 74.9% (from 0.24 to 0.06) and 58.9% (from 0.06 to 0.02) when the accuracy decreases by 0.46% (from 0.70 to 0.69) and 9.6% (from 0.67 to 0.60) respectively for the testing set. The results show that the FAB approach can considerably reduce fairness loss with only a slight reduction in classification error. The results also show that FAB can not only achieve obvious fairness improvement for the case when the original fairness loss is large (more than 0.2), but also enhance the fairness score evidently when the original fairness loss is small (less than 0.1). In addition, as shown in Figs.1-4, as the increase of $\lambda$, the fairness loss of FAB largely decreases while the classification accuracy slightly deteriorates.

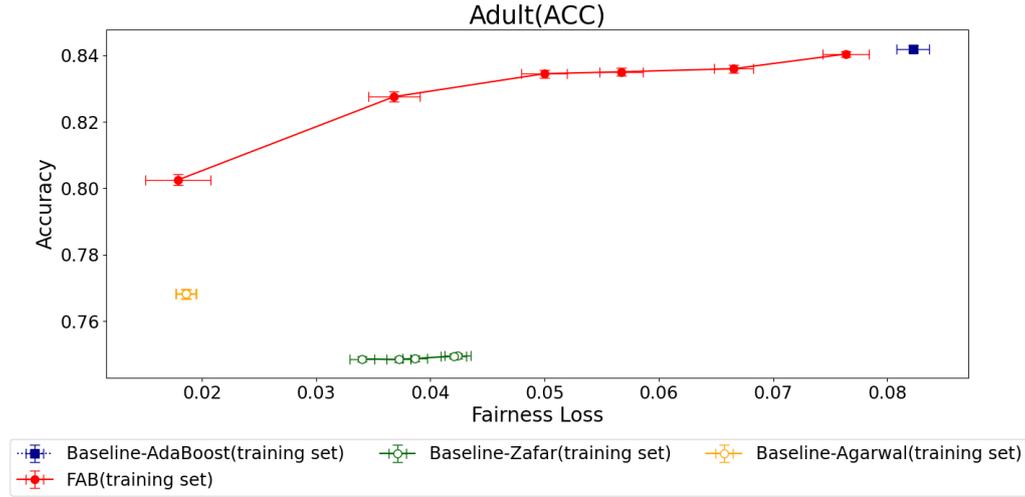

(a)

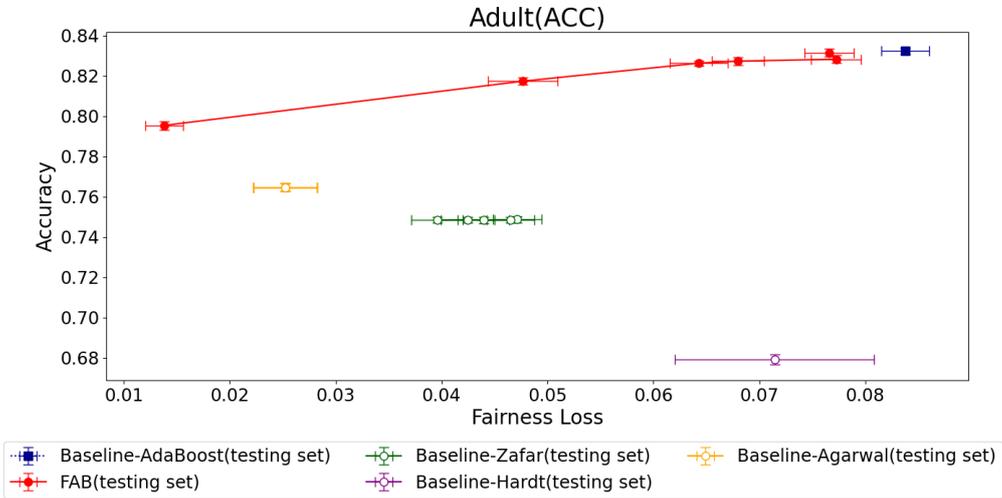

(b)

**Fig.1.** Classification accuracy and fairness indicator of accuracy of the proposed FAB approach and baselines on Adult dataset: (a) training performance, and (b) testing performance. For each plot, $\lambda$ increases from the upper right corner to the lower left corner.

Compared with Hardt, when considering accuracy / FPR for Adult, the accuracy of FAB is 17.1% (from 0.68 to 0.80) / 19.7% (from 0.68 to 0.81) higher than that of Hardt and the fairness loss of FAB is 80.7% (from 0.07 to 0.01) / 48.7% (from 0.05 to 0.02) less than that of Hardt. In terms of FNR for HSLS, when the fairness loss is similar (about 0.03), FAB can achieve 4.8% higher accuracy. Compared with Agarwal, when considering accuracy for Adult, the accuracy of FAB is 4.0% (from 0.76 to 0.80) higher than that of Agarwal and the fairness loss of FAB is 45.2% (from 0.03 to 0.01) less than that of Agarwal. In terms of FPR for Adult, when the fairness loss is similar (about 0.02), FAB can achieve 6.5% (from 0.76 to 0.81) higher accuracy. When considering FNR for HSLS, the accuracy and fairness loss are both similar.

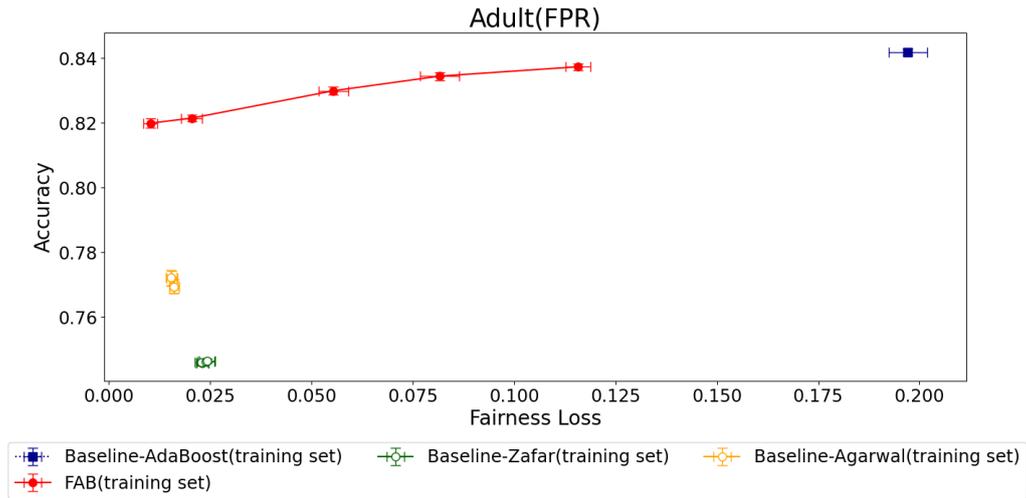

(a)

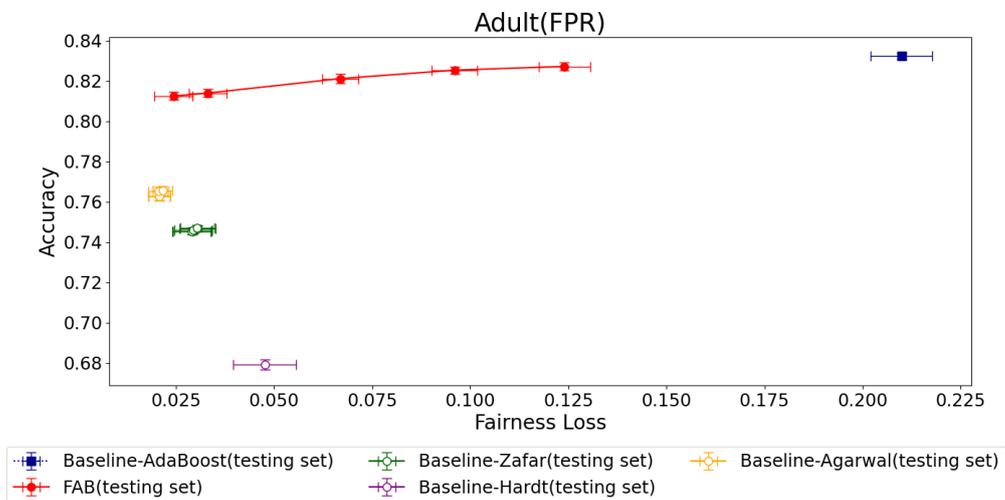

(b)

**Fig.2.** Classification accuracy and fairness indicator of FPR of the proposed FAB approach and baselines on Adult dataset: (a) training performance, and (b) testing performance. For each plot, $\lambda$ increases from the upper right corner to the lower left corner.

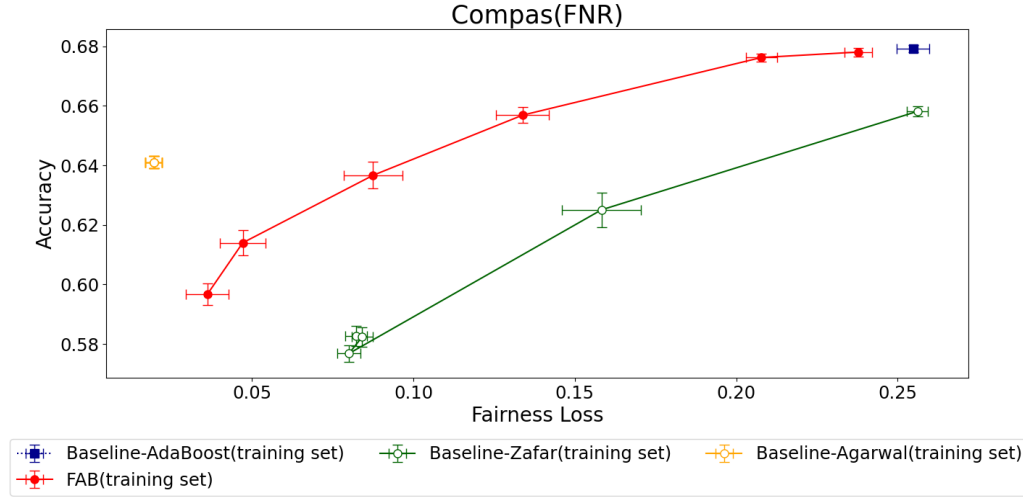

(a)

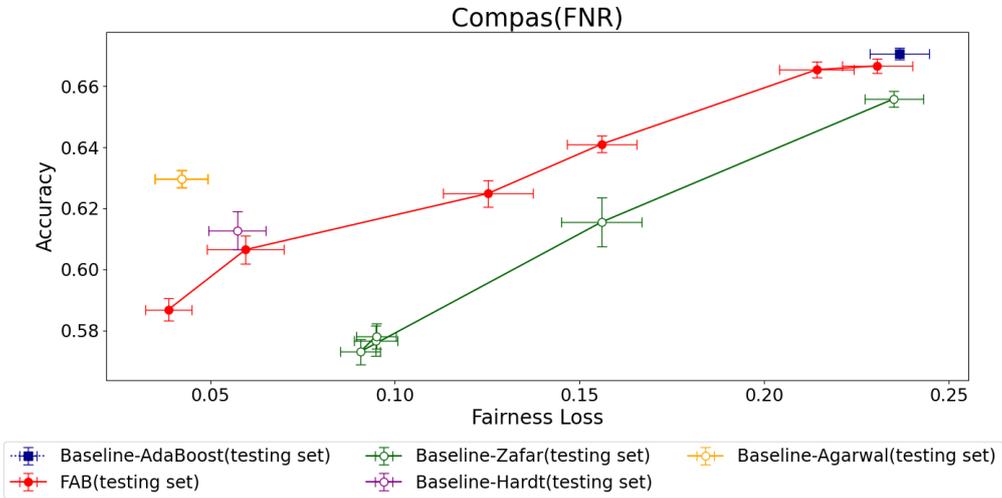

(b)

**Fig.3.** Classification accuracy and fairness indicator of FNR of the proposed FAB approach and baselines on Compas dataset: (a) training performance, and (b) testing performance. For each plot, $\lambda$ increases from the upper right corner to the lower left corner.

Compared with Zafar, in terms of accuracy / FPR for Adult, FAB can achieve 6.4% (from 0.75 to 0.80) / 9.0% (from 0.74 to 0.81) higher accuracy and 65.2% (from 0.04 to 0.01) / 16.4% (from 0.03 to 0.02) less fairness loss for the testing set. When it comes to FNR for Compas, the accuracy of the proposed FAB method is 5.2% higher than that of Zafar (from 0.58 to 0.61) and the fairness loss of FAB is 37.0% less than that of Zafar (from 0.09 to 0.06). Particularly, for Compas, though our method reduces fairness loss at greater cost of accuracy than other datasets, it still performs better than Zafar. When considering FNR for HSLS, Fig.4b shows that FAB can achieve higher classification accuracy when fairness loss is similar.

In summary, the proposed FAB method has superior performance in resolving the trade-off between classification accuracy and the fairness of a specific indicator.

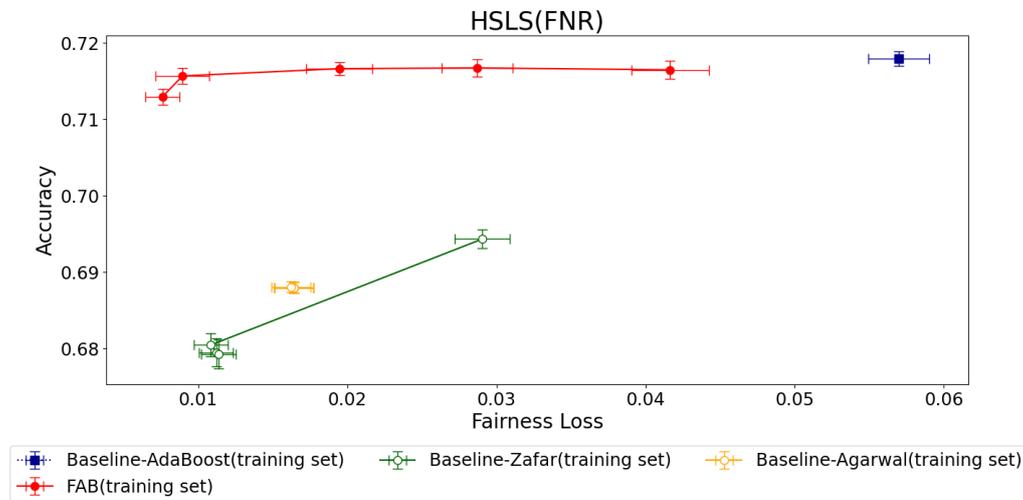

(a)

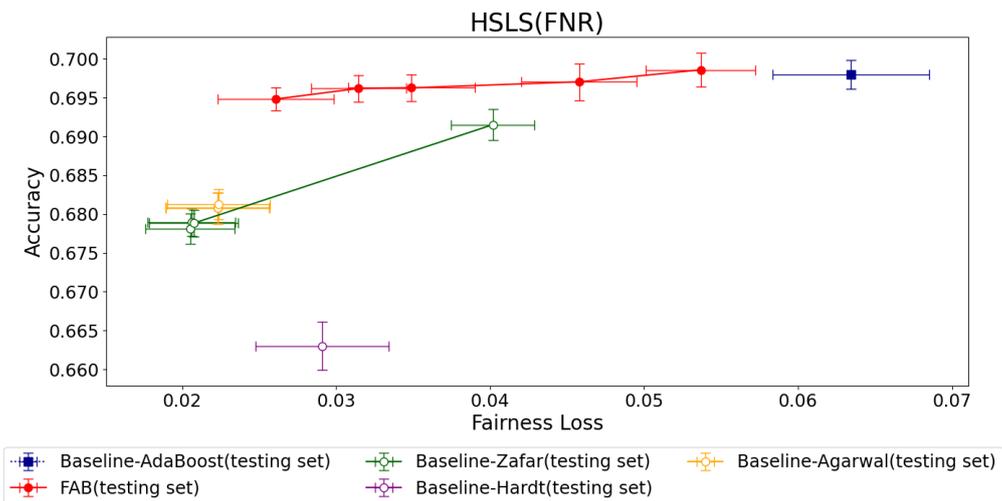

(b)

**Fig.4.** Classification accuracy and fairness indicator of FNR of the proposed FAB approach and baselines on HSLS dataset: (a) training performance, and (b) testing performance. For each plot, $\lambda$ increases from the upper right corner to the lower left corner.

**Table 2.** Representative experimental results including classification accuracy and fairness loss for testing set. For Fig.1b, we choose $\lambda = 0.5$ for FAB, $\varepsilon = 0.001$ for Agarwal and $\tau = 100$ for Zafar; for Fig. 2b, we choose $\lambda = 0.3$ for FAB, $\varepsilon = 0.001$ for Agarwal and $\tau = 0.01$ for Zafar; for Fig. 3b, we choose $\lambda = 0.4$ (the second point from the bottom left corner) for FAB, $\varepsilon = 0.001$ for Agarwal and $\tau = 0.01$ for Zafar; For Fig4b, we choose $\lambda = 0.25$ for FAB, $\varepsilon = 0.001$ for Agarwal and $\tau = 0.1$ for Zafar. Each experiment was repeated 20 times and the data here are the average of 20 experiments.

| Dataset | Fairness indicator | AdaBoost | | FAB | | Hardt | | Agarwal | | Zafar | |
|---|---|---|---|---|---|---|---|---|---|---|---|
| | | Accuracy | Fairness loss | Accuracy | Fairness loss | Accuracy | Fairness loss | Accuracy | Fairness loss | Accuracy | Fairness loss |
| Adult | Accuracy | 0.83 | 0.084 | 0.80 | 0.014 | 0.68 | 0.071 | 0.76 | 0.025 | 0.75 | 0.040 |
| Adult | FPR | 0.83 | 0.210 | 0.81 | 0.024 | 0.68 | 0.048 | 0.76 | 0.021 | 0.74 | 0.029 |
| Compas | FNR | 0.67 | 0.237 | 0.61 | 0.060 | 0.61 | 0.057 | 0.63 | 0.042 | 0.58 | 0.095 |
| HSLS | FNR | 0.70 | 0.064 | 0.69 | 0.026 | 0.66 | 0.029 | 0.68 | 0.022 | 0.68 | 0.021 |

## 5. Conclusions

In this article, a fairness-improving variant of AdaBoost named as FAB is proposed to achieve fair classification. A fairness-aware target loss function is adopted to quantify the error rate as well as unfairness. A hyperparameter $\lambda$ is introduced to show preferences for the fairness-accuracy trade-off, and a reweighting algorithm for base classifiers is derived accordingly. An upper bound for the target loss function of fairness-accuracy is theoretically derived, and an efficient computational algorithm that aims to minimize its upper bound is developed according to greedy strategy.

To evaluate the performance of FAB, experiments are conducted on three real-world datasets (i.e., Adult, COMPAS and HSLS) with respect to three fairness indicators of accuracy, FPR and FNR. The performance of FAB is quantified in comparison with four baselines in terms of classification accuracy and fairness loss. The results show that as the increase of $\lambda$, the fairness performance of FAB enhances significantly while its accuracy decreases slightly. It's also observed that FAB outperforms basic AdaBoost for a wide range of fairness-accuracy trade-offs, which significantly improves fairness at almost negligible accuracy decreases. With respect to four different settings (i.e., the fairness of accuracy for Adult, FPR for Adult, FNR for Compas, and FNR for HSLS), FAB can reduce the fairness loss by 83.5%, 88.4%, 74.9% and 58.9% at the cost of sacrificing the accuracy by 4.4%, 2.4%, 0.5% and 9.6% respectively. Moreover, FAB can not only achieve obvious fairness improvement when the original fairness loss is large (more than 0.2), but also enhance the fairness score when the original fairness loss is small (less than 0.1). When compared with the state-of-the-art fair classification methods proposed by Zafar, Hardt, and Agarwal, the proposed FAB approach also possesses superior performance in the fairness-accuracy trade-off.

**Appendix A. Derivation of upper bounds for the fairness of FPR and FNR**

In this section, the detailed derivation of the upper bounds for the fairness of FPR and FNR is provided.

*A.1. Upper bound for the fairness of FPR*

Given the assumption mentioned in problem statement (i.e., for general classifiers, there is $P(f(x) \neq y | S = 1, y = -1) < P(f(x) \neq y | S = 0, y = -1))$, Eq. (8) can be rewritten as:

$$L_{FPR}(f(x)) = \frac{1}{N}\sum_{i=1}^{N} \mathbb{I}(f(x_i) \neq y_i) + \lambda \left( \frac{\sum_{i=1}^{N} \mathbb{I}(S_i=0)\mathbb{I}(y_i=-1)\mathbb{I}(f(x_i)\neq y_i)}{\sum_{i=1}^{N} \mathbb{I}(S_i=0)\mathbb{I}(y_i=-1)} - \frac{\sum_{i=1}^{N} \mathbb{I}(S_i=1)\mathbb{I}(y_i=-1)\mathbb{I}(f(x_i)\neq y_i)}{\sum_{i=1}^{N} \mathbb{I}(S_i=1)\mathbb{I}(y_i=-1)} \right) = \sum_{i=1}^{N} \mathbb{I}(S_i=0)\mathbb{I}(y_i=-1)\left( \frac{1}{N} + \frac{\lambda}{\sum_{i=1}^{N} \mathbb{I}(S_i=0)\mathbb{I}(y_i=-1)} \right)\mathbb{I}(f(x_i) \neq y_i) + \sum_{i=1}^{N} \mathbb{I}(S_i=1)\mathbb{I}(y_i=-1)\left( \frac{1}{N} - \frac{\lambda}{\sum_{i=1}^{N} \mathbb{I}(S_i=1)\mathbb{I}(y_i=-1)} \right)\mathbb{I}(f(x_i) \neq y_i) + \sum_{i=1}^{N} \mathbb{I}(y_i = +1)\frac{1}{N}\mathbb{I}(f(x_i) \neq y_i) \quad (23)$$

Combining this result with Eq. (1), Eq. (9), Eq. (14) and Eq. (16), when $\lambda \in \left[ 0, \frac{\sum_{i=1}^{N} \mathbb{I}(S_i=1)\mathbb{I}(y_i=-1)}{N} \right]$, we can obtain:

$$L_{FPR}(f(x)) \leq \sum_{i=1}^{N} \mathbb{I}(S_i = 0)\mathbb{I}(y_i = -1)\left(\frac{1}{N} + \frac{\lambda}{\sum_{i=1}^{N} \mathbb{I}(S_i=0)\mathbb{I}(y_i=-1)}\right)e^{-y_i \sum_{t=1}^{T} \alpha_t h_t(x_i)} + \sum_{i=1}^{N} \mathbb{I}(S_i = 1)\mathbb{I}(y_i = -1)\left(\frac{1}{N} - \frac{\lambda}{\sum_{i=1}^{N} \mathbb{I}(S_i=1)\mathbb{I}(y_i=-1)}\right)e^{-y_i \sum_{t=1}^{T} \alpha_t h_t(x_i)} + \sum_{i=1}^{N} \mathbb{I}(y_i = +1)\frac{1}{N}e^{-y_i \sum_{t=1}^{T} \alpha_t h_t(x_i)} = \sum_{i=1}^{N} D_1^{FPR}(x_i)e^{-y_i \sum_{t=1}^{T} \alpha_t h_t(x_i)} = \prod_{t=1}^{T} Z_t^{FPR}(h_t, \alpha_t)$$
(24)

*A.2. Upper bound for the fairness of FNR*

Given the assumption mentioned in problem statement (i.e., for general classifiers, there is $P(f(x) \neq y | S = 1, y = +1) < P(f(x) \neq y | S = 0, y = +1))$, Eq. (10) can be rewritten as:

$$L_{FNR}(f(x)) = \frac{1}{N}\sum_{i=1}^{N} \mathbb{I}(f(x_i) \neq y_i) + \lambda\left(\frac{\sum_{i=1}^{N} \mathbb{I}(S_i=0)\mathbb{I}(y_i=+1)\mathbb{I}(f(x_i) \neq y_i)}{\sum_{i=1}^{N} \mathbb{I}(S_i=0)\mathbb{I}(y_i=+1)} - \frac{\sum_{i=1}^{N} \mathbb{I}(S_i=1)\mathbb{I}(y_i=+1)\mathbb{I}(f(x_i) \neq y_i)}{\sum_{i=1}^{N} \mathbb{I}(S_i=1)\mathbb{I}(y_i=+1)}\right) = \sum_{i=1}^{N} \mathbb{I}(S_i = 0)\mathbb{I}(y_i = +1)\left(\frac{1}{N} + \frac{\lambda}{\sum_{i=1}^{N} \mathbb{I}(S_i=0)\mathbb{I}(y_i=+1)}\right)\mathbb{I}(f(x_i) \neq y_i) + \sum_{i=1}^{N} \mathbb{I}(S_i = 1)\mathbb{I}(y_i = +1)\left(\frac{1}{N} - \frac{\lambda}{\sum_{i=1}^{N} \mathbb{I}(S_i=1)\mathbb{I}(y_i=+1)}\right)\mathbb{I}(f(x_i) \neq y_i) + \sum_{i=1}^{N} \mathbb{I}(y_i = -1)\frac{1}{N}\mathbb{I}(f(x_i) \neq y_i)$$
(25)

Combining this result with Eq. (1), Eq. (11), Eq. (14) and Eq. (16), when $\lambda \in \left[0, \frac{\sum_{i=1}^{N} \mathbb{I}(S_i=1)\mathbb{I}(y_i=+1)}{N}\right]$, we have:

$$L_{FNR}(f(x)) \leq \sum_{i=1}^{N} \mathbb{I}(S_i = 0)\mathbb{I}(y_i = +1)\left(\frac{1}{N} + \frac{\lambda}{\sum_{i=1}^{N} \mathbb{I}(S_i=0)\mathbb{I}(y_i=+1)}\right)e^{-y_i \sum_{t=1}^{T} \alpha_t h_t(x_i)} + \sum_{i=1}^{N} \mathbb{I}(S_i = 1)\mathbb{I}(y_i = +1)\left(\frac{1}{N} - \frac{\lambda}{\sum_{i=1}^{N} \mathbb{I}(S_i=1)\mathbb{I}(y_i=+1)}\right)e^{-y_i \sum_{t=1}^{T} \alpha_t h_t(x_i)} + \sum_{i=1}^{N} \mathbb{I}(y_i = -1)\frac{1}{N}e^{-y_i \sum_{t=1}^{T} \alpha_t h_t(x_i)} = \sum_{i=1}^{N} D_1^{FNR}(x_i)e^{-y_i \sum_{t=1}^{T} \alpha_t h_t(x_i)} = \prod_{t=1}^{T} Z_t^{FNR}(h_t, \alpha_t)$$
(26)

**Appendix B.** A table summarizing the main variables used in this article is provided below.

Table 3. Main variables used in this work

| Variable | Meaning |
|---|---|
| $x_i$ | Features of the *i*th sample |
| $y_i$ | Binary true label of the *i*th sample |
| $S_i$ | Sensitive feature of the *i*th sample (belongs to $x_i$) |
| $\mathcal{D} = \{x_i, y_i\}_{i=1}^N$ | Training dataset |
| $\hat{y}_i$ | Predicted label of the *i*th sample |
| $h_t(x)$ | Base classifier of the *t*th iteration |
| $\alpha_t$ | Weight of $h_t(x)$ |
| $f(x)$ | Ensemble classifier |
| $\lambda$ | Hyperparameter for the trade-off between accuracy and fairness |
| $e_t$ | Weighted error rate of the *t*th iteration |
| $D_t(x_i)$ | Sample weight of the *i*th sample during the *t*th iteration |
| $Z_t(h_t, \alpha_t)$ | Normalization factor of $D_t$ |
| $D_t^{acc}(x_i)$ | Sample weight of the *i*th sample during the *t*th iteration in terms of the fairness of accuracy |
| $Z_t^{acc}(h_t, \alpha_t)$ | Normalization factor of $D_t^{acc}$ |
| $D_t^{FPR}(x_i)$ | Sample weight of the *i*th sample during the *t*th iteration in terms of the fairness of FPR |
| $Z_t^{FPR}(h_t, \alpha_t)$ | Normalization factor of $D_t^{FPR}$ |
| $D_t^{FNR}(x_i)$ | Sample weight of the *i*th sample during the *t*th iteration in terms of the fairness of FNR |
| $Z_t^{FNR}(h_t, \alpha_t)$ | Normalization factor of $D_t^{FNR}$ |
| $\mathbb{I}(\partial)$ | $\mathbb{I}(\partial) = \begin{cases} 1, & \text{if and only if } \partial \text{ is true} \\ 0, & \text{otherwise} \end{cases}$ |
| $L_{acc}(f(x))$ | Loss function for $f(x)$ in terms of the fairness of accuracy |
| $L_{FPR}(f(x))$ | Loss function for $f(x)$ in terms of the fairness of FPR |
| $L_{FNR}(f(x))$ | Loss function for $f(x)$ in terms of the fairness of FNR |

**CRediT authorship contribution statement**

Xiaobin Song: Conceptualization, Methodology, Software, Writing - original draft. Zeyuan Liu: Conceptualization, Writing - original draft. Benben Jiang: Methodology, Writing - review & editing.

**Declaration of competing interest**

The authors declare that they have no known competing financial interests or personal relationships that could have appeared to influence the work reported in this work.

**Acknowledgments**

This work is supported by the National Key Research and Development Program of China (2022YFE0197600) and the National Natural Science Foundation of China (62273197).